\documentclass{article}
\PassOptionsToPackage{numbers, compress}{natbib}
\usepackage[preprint]{neurips_2024}

\usepackage[utf8]{inputenc} 
\usepackage[T1]{fontenc}    
\usepackage{hyperref}       
\usepackage{url}            
\usepackage{booktabs}       
\usepackage{amsfonts}       
\usepackage{nicefrac}       
\usepackage{microtype}      
\usepackage{xcolor}         
\usepackage{amsmath}
\usepackage{graphicx}
\usepackage{wrapfig}
\usepackage{adjustbox}
\usepackage{multirow}
\usepackage{algorithm}
\usepackage{algorithmic}

\title{Magnitude-based Neuron Pruning for Backdoor Defense}

\author{%
  Nan Li\\
  School of Cyber Science and Engineering\\
  Shanghai Jiao Tong University\\
  Shanghai, China, 200240\\
  \texttt{lyyqmjshu@sjtu.edu.cn} \\
  \And
  Haoyu Jiang\\
  School of Cyber Science and Engineering\\
  Shanghai Jiao Tong University\\
  Shanghai, China, 200240\\
  \texttt{jhy549@sjtu.edu.cn} \\
  \AND
  Ping Yi\\
  School of Cyber Science and Engineering\\
  Shanghai Jiao Tong University\\
  Shanghai, China, 200240\\
  \texttt{yiping@sjtu.edu.cn} \\
}

\begin{document}

\maketitle

\begin{abstract} \label{abstract}
  Deep Neural Networks (DNNs) are known to be vulnerable to backdoor attacks, posing concerning threats to their reliable deployment. Recent research reveals that backdoors can be erased from infected DNNs by pruning a specific group of neurons, while how to effectively identify and remove these backdoor-associated neurons remains an open challenge. In this paper, we investigate the correlation between backdoor behavior and neuron magnitude, and find that backdoor neurons deviate from the magnitude-saliency correlation of the model. The deviation inspires us to propose a Magnitude-based Neuron Pruning (MNP) method to detect and prune backdoor neurons. Specifically, 
  MNP uses three magnitude-guided objective functions to manipulate the magnitude-saliency correlation of backdoor neurons, thus achieving the purpose of exposing backdoor behavior, eliminating backdoor neurons and preserving clean neurons, respectively. Experiments show our pruning strategy achieves state-of-the-art backdoor defense performance against a variety of backdoor attacks with a limited amount of clean data, demonstrating the crucial role of magnitude for guiding backdoor defenses.
\end{abstract}

\section{Introduction} \label{introduction}
In recent years, Deep Neural Networks (DNNs) have demonstrated remarkable capabilities in solving real-world problems. However, the wide application of DNNs has raised concerns about their security and trustworthiness. Recent works have shown that DNNs are vulnerable to backdoor attacks\cite{guBadnetsIdentifyingVulnerabilities2017}, in which an adversary injects malicious triggers into the victim model through data poisoning, manipulating the training process, or directly modifying model parameters. The backdoored model performs well on clean samples but can be triggered into false predictions by the poisoned samples containing trigger patterns. As pre-trained weights and outsourced training are widely applied to cut computational costs for training DNNs, the backdoor attack is becoming an undeniable security issue. 

To address this issue, numerous methods have been proposed for detecting and mitigating backdoor attacks. Backdoor detection methods \cite{wangNeuralCleanseIdentifying2019,liuABSScanningNeural2019,huTriggerHuntingTopological2021} identify whether a model is backdoored or a dataset is poisoned, while backdoor mitigation methods \cite{liuFinepruningDefendingBackdooring2018,liNeuralAttentionDistillation2020} eliminate the injected triggers from backdoored models. Recent research \cite{wuAdversarialNeuronPruning2021,liReconstructiveNeuronPruning2023} has observed a subset of neurons contributing the most to backdoor behaviors in infected DNNs. By pruning these backdoor-associated neurons, the backdoor behavior of the infected model can be effectively mitigated. Backdoor neurons are believed to have certain properties. For example, they can only be activated by trigger patterns \cite{zhuEnhancingFineTuningBased2023}, and are more sensitive to input \cite{zhengDatafreeBackdoorRemoval2022} or perturbation \cite{wuAdversarialNeuronPruning2021}. These properties can be used to design certain pruning strategies to mitigate the injected backdoor.

Magnitude is considered an important indicator to guide pruning in model compression studies \cite{liPruningFiltersEfficient2016,heSoftFilterPruning2018}. Most of these studies assume a positive correlation between neuron magnitude and neuron importance for model performance, as neurons with smaller magnitudes have less numerical impact on the output of the model. We empirically show that backdoor neurons deviate from this correlation, as they contain extra weights used to trigger the backdoor. Similar observations are also implied in \cite{zhengDatafreeBackdoorRemoval2022,zhuEnhancingFineTuningBased2023}. Motivated by our findings, we propose a Magnitude-based Neuron Pruning (MNP) method to defend against backdoor attacks. Specifically, MNP first detects the injected backdoor by analyzing the correlation between neuron magnitude and neuron saliency, and then optimizes neuron masks with three magnitude-guided objective functions to expose and prune backdoor neurons. Given a small subset of clean samples, MNP can effectively defend against backdoors injected by a variety of backdoor attacks. Experiments show MNP is competitive for both backdoor detection and mitigation among a set of state-of-the-art backdoor defense methods. Our defense strategy significantly surpasses the previous state-of-the-art method RNP \cite{liReconstructiveNeuronPruning2023}, achieving superior defense outcomes against over ten types of attacks on both the CIFAR-10 and ImageNet datasets, with the majority of results reaching optimal performance levels.

To summarize, our main contributions are three-fold. 
\textbf{1)} We explored the correlation between neuron magnitude and their contribution to backdoor behavior, and try to reinterpret the mechanism of backdoor defenses from the perspective of neuron magnitude.
\textbf{2)} Beyond offering hypothesis for how neuron magnitude works in backdoor defense methods, we further validate our claims by using them as principles to construct our own backdoor defense method MNP, which utilize three optimization objectives to manipulate the magnitude of neurons and 
\textbf{3)} We empirically show that our method is competitive compared to ten state-of-the-art backdoor defense methods against ten challenging backdoor attacks across different model architectures and datasets.

\section{Related Work}\label{rw}
\paragraph{Backdoor Attack.} Depending on how the trigger pattern is injected, backdoor attacks fall into two main categories: input-space attacks poisoning the training dataset and feature-space attacks manipulating the training process or directly modifying model parameters. Input-space attacks, also known as poisoning-based attacks, are conducted by modifying a small subset of training data, which commonly includes patching trigger patterns into the sample and shifting the corresponding label to the targeted class. Note that there are also clean-label attacks \cite{turnerCleanlabelBackdoorAttacks2019} that do not relabel any sample. The model trained on the poisoned data learns both the clean task permitting its accuracy on clean samples, and the backdoor task tricking it into false predictions with the trigger pattern. The input-space attacks can be further categorized into static and dynamic attacks depending on the trigger pattern they use. Static attacks use the same trigger pattern for all samples, such as black-white squares \cite{guBadnetsIdentifyingVulnerabilities2017}, Gaussian noise \cite{chenTargetedBackdoorAttacks2017} and adversarial perturbations \cite{turnerCleanlabelBackdoorAttacks2019}, while dynamic attacks \cite{nguyenInputAwareDynamicBackdoor2020,nguyenWaNetImperceptibleWarpingbased2021a} use sample-wise triggers to enhance their stealthiness. Feature-space attacks occur under a different threat model, where the adversary has full access to the training process and the model weights. These attacks may directly manipulate the training process with certain optimization objectives \cite{shafahiPoisonFrogsTargeted2018,cheng2021deep,zhaoDEFEATDeepHidden2022}, or perturb the model weights \cite{gargCanAdversarialWeight2020,qiPracticalDeploymentStageBackdoor2022} to inject backdoor in the feature space, making them more challenging to defend against.

\paragraph{Backdoor Defense.} Backdoor defense involves two primary tasks: backdoor detection and backdoor mitigation. Backdoor detection methods focus on identifying backdoored models \cite{barni2019new,liuABSScanningNeural2019,chenDeepinspectBlackboxTrojan2019, kolouriUniversalLitmusPatterns2020a} or backdoored samples \cite{gaoSTRIPDefenceTrojan2020, DBLP:conf/nips/Tran0M18}. Some advanced detection methods also conduct reverse engineering with the backdoored model to recover the trigger pattern \cite{wangNeuralCleanseIdentifying2019, taoBetterTriggerInversion2022, wangRethinkingReverseengineeringTrojan2022, huTriggerHuntingTopological2021}. Backdoor mitigation methods aim to remove the injected backdoor from the infected model with minimal degradation of its performance on clean samples. Existing techniques include fine-tuning, distillation \cite{liNeuralAttentionDistillation2020}, unlearning \cite{zengAdversarialUnlearningBackdoors2022}, pruning \cite{liuFinepruningDefendingBackdooring2018}, and training-time defenses \cite{liAntiBackdoorLearningTraining2021,wangTrainingMoreConfidence2022,liuABSScanningNeural2019}. Recent works on pruning have demonstrated remarkable performance in backdoor mitigation. FP \cite{liuFinepruningDefendingBackdooring2018} assumes backdoor-associated neurons can only be activated by the trigger pattern, thus pruning neurons that are dominant when feeding the model clean samples can promisingly mitigate the injected backdoor. ANP \cite{wuAdversarialNeuronPruning2021} perturbs neuron weights to maximize the classification loss of the model, then fixes the model by pruning neurons that are more sensitive to the adversarial perturbation, as these neurons are believed to be strongly related to the injected backdoor. RNP \cite{liReconstructiveNeuronPruning2023} optimizes masks through an unlearning-recovering process to expose backdoor neurons. CLP \cite{zhengDatafreeBackdoorRemoval2022} introduces the channel lipschitz value to evaluate each neuron's sensitivity to input and prunes backdoor neurons with high sensitivity. FT-SAM \cite{zhuEnhancingFineTuningBased2023} has revealed that backdoor neurons tends to have larger magnitudes, and incorporate sharpness-aware minimization with fine-tuning to purify the injected models. 

\section{Preliminaries}\label{sect3}
\subsection{Backdoor Learning}\label{bl}
Consider a standard $K$-class classification problem on a training set $\mathcal{D} = \{(\boldsymbol{x}_i, y_i)\}^D_{i=1}\subseteq \mathcal{X}\times\mathcal{Y}$, with $\mathcal{X}\subset \mathbb{R}^d$ as the sample space and $\mathcal{Y} \subset \{1, 2,...,K\}$ as the label space. Given a subset $\mathcal{D}_b\subseteq\mathcal{D}$, the standard poisoning-based backdoor attack involves injecting the trigger pattern into input samples with the poisoning function $\delta:\mathcal{X}\rightarrow \mathcal{X}$ and modifying corresponding labels with the label shifting function $S:\mathcal{Y}\rightarrow\mathcal{Y}$. The backdoor attack can be viewed as a multi-task learning problem \cite{liReconstructiveNeuronPruning2023} on both the clean subset $\mathcal{D}_c = \mathcal{D} - \mathcal{D}_b$ and the backdoor subset $\mathcal{D}_b$. Let $F$ denote the victim model with parameter $\theta$. We consider a parametric hypothesis that the parameter space of $F$ can be decomposed into $\theta = \theta_c\cup\theta_b\cup \theta_{sh}$, where $\theta_c$, $\theta_b$ and $\theta_{sh}$ denote the clean, the backdoor and the shared neurons, respectively. We further assume that $\theta_{sh}$ can be omitted since the backdoor and the clean tasks are highly independent, i.e., backdoor attacks are designed not to affect the model performance on clean samples \cite{guBadnetsIdentifyingVulnerabilities2017}. The standard backdoor learning process can be expressed as follows:
\begin{equation}
\underset{\theta=\theta_c \cup \theta_b}{\arg \min } \big[{\underbrace{\mathbb{E}_{\left(\boldsymbol{x}, y\right) \in \mathcal{D}_c} \mathcal{L}\left(F\left(\boldsymbol{x}; \theta_c\right), y\right)}_{\textrm{clean task}} +\underbrace{\mathbb{E}_{\left(\boldsymbol{x}, y\right) \in \mathcal{D}_b} \mathcal{L}\left(F\left(\delta(\boldsymbol{x}); \theta_b\right), S(y)\right)}_{\textrm{backdoor task}}}\big],
\end{equation}

\subsection{Neuron Magnitude and Saliency}
Consider a convolutional network $F$ with $L$ layers, regarding the fully connected layer as the convolutional layer with $1\times 1$ kernels. Let $f^{(i)}$ denote the $i$-th layer with the weight matrix $\theta^{(i)}\in \mathbb{R}^{c_{i}\times c_{i-1}\times h\times w}$, where $c_i$, $h$ and $w$ denote the channel number of $f^{(i)}$, the height and the width of the convolutional kernel, respectively. $\theta^{(i)}$ consists of $c_i$ filters $\{\theta^{(i,j)}\in \mathbb{R}^{c_{i-1}\times h\times w}\}_{j=1}^{c_i}$. Pruning the $j$-th filter of the $i$-th layer refers to setting $\theta^{(i,j)}$ to an all-zero matrix, thus removing the corresponding output feature map. 
We denote the $l_p$-norm of a filter as $\Vert \theta^{(i,j)}\Vert_p$, which is widely used to evaluate the magnitude of each filter in model compression methods \cite{liPruningFiltersEfficient2016,heSoftFilterPruning2018}. 
Recent research \cite{zhuEnhancingFineTuningBased2023} has observed a strong positive correlation between backdoor behavior and the weight norms for each neuron, which implies that backdoor neurons may have larger magnitudes than clean ones. To further quantify the contribution of each neuron to the backdoor and the clean task, we introduce the concept of neuron saliency, which is commonly defined as the loss change induced by pruning that neuron in model compression research \cite{lecunOptimalBrainDamage1989}. Given a test set $\mathcal{D}_t$, for each filter, we define the saliency metrics \textit{Clean Loss Change} (CLC) and \textit{Backdoor Loss Change} (BLC) as follows:
\begin{align}
\label{clc}\textrm{CLC}(\theta,i,j) &= \mathbb{E}_{(\boldsymbol{x}, y) \in \mathcal{D}_t}\mathcal{L}(F(\boldsymbol{x}; \theta|{\theta^{(i,j)}=0}), y) -\mathbb{E}_{(\boldsymbol{x}, y) \in \mathcal{D}_t}\mathcal{L}(F(\boldsymbol{x}; \theta), y),\\
\textrm{BLC}(\theta,i,j) &= \mathbb{E}_{(\boldsymbol{x}, y) \in \mathcal{D}_t}\mathcal{L}(F(\delta(\boldsymbol{x}); \theta|{\theta^{(i,j)}=0}), S(y)) -\mathbb{E}_{(\boldsymbol{x}, y) \in \mathcal{D}_t}\mathcal{L}(F(\delta(\boldsymbol{x}); \theta), S(y))
\end{align}

\begin{figure}[t]
\centering
\includegraphics[width = \textwidth]{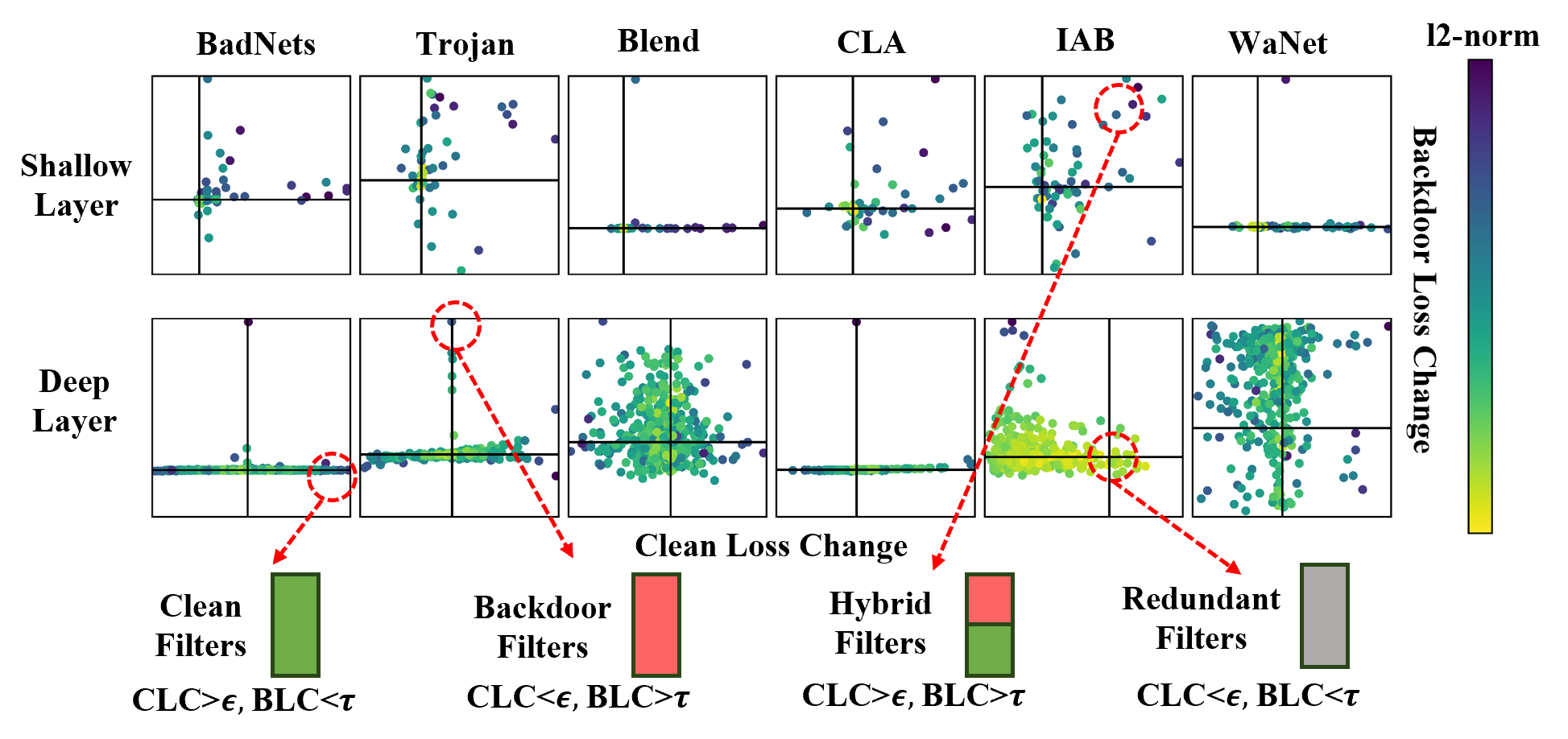}
\caption{Scatter plots depicting the BLC and CLC of filters in the shallow and deep convolutional layers of backdoored ResNet18 models, attacked by \textbf{BadNets}, \textbf{Trojan}, \textbf{Blend}, \textbf{CLA}, \textbf{IAB}, and \textbf{WaNet}. Quadrants are determined by the horizontal and vertical lines (x-axis and y-axis) at $\textrm{CLC}=0$ and $\textrm{BLC}=0$. The color of each point indicates the $l_2$-norm of the corresponding filter weight, with deeper colors representing larger $l_2$-norms.}
\label{dis}
\end{figure}

\subsection{Observations about Backdoor and Clean Neurons}\label{obs}
We investigate the distribution of the CLC, BLC and magnitude of filters across different layers of backdoored models, as shown in Fig \ref{dis}. Our key observations are as follows:


\paragraph{Backdoor and clean neurons may overlap.}
Most of the recent works \cite{liuFinepruningDefendingBackdooring2018,zhengPreactivationDistributionsExpose2022} assume that backdoor and clean neurons hardly overlap in filter level, i.e., each filter is associated with either the backdoor task or the clean task. However, filters in the first quadrants in Fig \ref{dis} have positive BLC and positive CLC, which implies they contribute to both clean accuracy and backdoor behavior. Instead of simply defining a filter as backdoored or clean, we suggest a fine-grained approach with thresholds $\tau\geq0, \epsilon\geq0$ to categorize filters into backdoor, clean, hybrid and redundant filters. We follow the assumption mentioned in Section \ref{bl}, that the backdoor and clean parameters are separated at the neuron level. A filter is composed of a large number of neurons, as it commonly contains multiple channels and kernel weights. Backdoor filters primarily consist of backdoor neurons that are crucial for the backdoor task but have minimal contribution to the clean task, while clean filters exhibit the opposite behavior. Hybrid filters are composed of both backdoor neurons and clean neurons, hence, pruning them results in an increase in both backdoor loss and clean loss. Redundant filters contain unimportant neurons and can be pruned without significantly affecting the overall performance of the model, as is widely observed in model compression studies \cite{liPruningFiltersEfficient2016,heSoftFilterPruning2018}.

\paragraph{The correlation between neuron magnitude and saliency.}
As weight decay is widely applied to reduce the complexity of DNNs, neurons of less importance tend to have smaller magnitudes. Based on the above assumption, magnitude-based model compression methods \cite{liPruningFiltersEfficient2016,heSoftFilterPruning2018} use the $l_p$-norm of filters as a statistical indicator of their contribution to the final prediction result of the model
. Since CLC measures the importance of filters for model performance, we assume a positive correlation between neuron magnitude and the CLC in the clean parameter space $\theta_c$. Denoting the correlation by $C:\mathbb{R}\rightarrow\mathbb{R}$, every clean filter $\theta^{(i,j)}$ satisfies $\textrm{CLC}(F,i,j)=C(\Vert\theta^{(i,j)}\Vert_p)$. Since the backdoor model is trained on both the backdoor task and the clean task, each filter $\theta^{(i,j)}$ can be decomposed into $\theta^{(i,j)} = \theta^{(i,j)}_c \cup \theta^{(i,j)}_b$. Its contribution to the clean task satisfies:
\begin{equation}
\textrm{CLC}(F, i,j) = C(\Vert\theta^{(i,j)}_{c}\Vert_p)\leq C(\Vert\theta^{(i,j)}\Vert_p),
\end{equation}
which indicates that the $l_p$-norm of the backdoor or hybrid filter does not correspond to its actual contribution to the clean task, for it contains additional parameters used to trigger the backdoor. This observation is consistent with the results in Fig \ref{dis}, where the majority of backdoor filters have larger $l_p$-norms compared to clean filters with the same CLC value. 

\section{Methodology}
\subsection{Basic settings}\label{ass}
\paragraph{Assumptions.}
Our research is based on two fundamental assumptions about the neurons of the backdoored DNN: \textbf{1)} A filter can be associated with both the backdoor task and the clean task, if so, it is composed of both clean and backdoor neurons. \textbf{2)} The model is trained with weight decay or other parameter regularization techniques, thus the $l_p$ norm of a filter is positively correlated with its overall contribution to clean accuracy and backdoor behavior.
\paragraph{Defense setting.}
We adopt a typical defense setting where the defender has downloaded a backdoored model from an untrustworthy third party without knowledge of the attack or training data. We assume a small amount of clean data $\mathcal{D}_d$ is available for defense, which can be collected from the Internet or carefully selected from the training data. The defender aims to first determine if the model is backdoored, then remove the backdoor behavior from the infected model with minimum degradation to its clean accuracy.

\subsection{Magnitude-guided Optimization} \label{sec4.2}
A number of recent works \cite{wuAdversarialNeuronPruning2021,liReconstructiveNeuronPruning2023,chaiOneshotNeuralBackdoor2022,zhuEnhancingFineTuningBased2023} on backdoor mitigation have employed a min-max optimization process to expose backdoor neurons. However, most of these works have not explicitly included the magnitude of neurons in their optimization objectives. More analyses can be found in \ref{minmax}. Based on our analysis on how neuron magnitude works in backdoor defense approaches, we consider three optimization objectives to control the magnitude of neurons: weight penalty, clean suppression and clean preserving.
\paragraph{Weight Penalty.}
As is discussed in Section \ref{obs}, the $l_p$-norms of clean filters approximately follow a positive correlation with their contribution to clean accuracy, while backdoor or hybrid filters contain backdoor neurons and have larger $l_p$-norms that deviate from the positive correlation. 
The deviation inspires us to develop a strategy to prune filters with large $l_p$-norms while preserving (or recovering) the clean accuracy.
For each filter, we apply a mask $m \in [0,1]$, which acts on the magnitude of each filter without changing the specific weight. 
For model $F$ with $L$ layers, the collection of all masks is denoted by $\mathcal{M} = \{\boldsymbol{m}_i\}_{i=1}^L$, where $\boldsymbol{m}_i\in [0,1]^{c_i}$ is a vector of masks for the $i$-th layer with $c_i$ filters. 
The masked network $F_{\mathcal{M}}$ has the same architecture as $F$ with weight matrices of all the convolutional layers set to $\boldsymbol{m}_i\odot\theta^{(i)}$. To expose the backdoor filters with large $l_p$-norms and minimal impact on the clean loss, we formulate our problem as follows:
\begin{equation}
    \min_{\boldsymbol{m}_i\in [0,1]^{c_i}} \big[\mathbb{E}_{(\boldsymbol{x}, y) \in \mathcal{D}_d}\mathcal{L}(F_{\mathcal{M}}(\boldsymbol{x}; \theta), y)
    + \lambda \sum_{i=1}^L\Vert\boldsymbol{w}_i\odot \boldsymbol{m}_i\Vert_1\big],
\end{equation}
where $\odot$ denotes the Hadamard product, $\boldsymbol{w}_i\in \mathbb{R}^{c_i}$ is the vector of $l_2$-norms of filters of the $i$-th convolutional layer, and $\lambda>0$ is a hyperparameter balancing the loss and the weight penalty term.
\paragraph{Clean Suppression.}
The clean suppression objective is designed to reduce the magnitude of most clean neurons and expose the backdoor behavior of the infected model:
\begin{equation}
\max_{\theta} \big[\mathbb{E}_{(\boldsymbol{x}, y) \in \mathcal{D}_d}\mathcal{L}(F(\boldsymbol{x}; \theta), y) - \mu \Vert\theta\Vert_2^2\big],
\end{equation}
where $\mu>0$ is the weight decay hyperparameter. 
Since the suppression of clean neurons both increases the classification loss and reduces the $l_2$ regularization term, during the optimization, magnitude of clean neurons decrease at a faster rate than backdoor and hybrid neurons.
\paragraph{Clean Preserving.}
In contrast to clean suppression, the clean preservation objective is designed to preserve an important subset of clean neurons. 
This includes most clean filters and part of hybrid filters that are critical for the clean accuracy. 
Similarly to the weight penalty process, we initialize $\boldsymbol{m}_i\in [1,2]^{c_i}$ for each filter and optimize masks to increase magnitude of filters as much as possible while retaining the clean loss:
\begin{equation}
  \min_{\boldsymbol{m}_i\in [1,2]^{c_i}} \big[\mathbb{E}_{(\boldsymbol{x}, y) \in \mathcal{D}_d}\mathcal{L}(F_{\mathcal{M}}(\boldsymbol{x}; \theta), y)
  - \lambda \sum_{i=1}^L\Vert\boldsymbol{w}_i\odot \boldsymbol{m}_i\Vert_1\big],
\end{equation}

\begin{figure}[t]
\centering
\includegraphics[width = 0.95\textwidth]{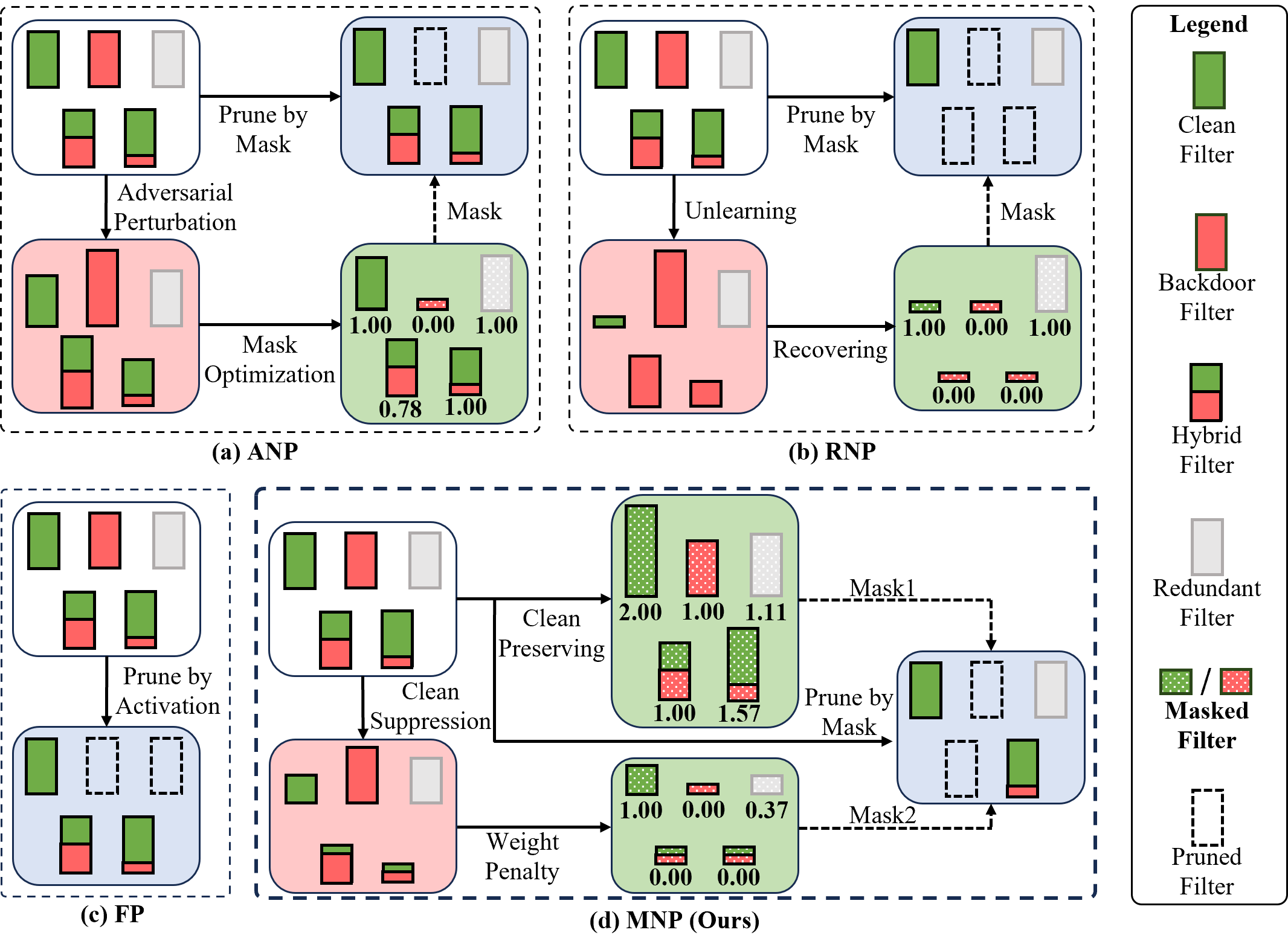}
\caption{Overview of our proposed \textbf{MNP} framework, in comparison with 3 existing backdoor mitigation methods: \textbf{ANP}, \textbf{RNP}, and \textbf{FP}. MNP exposes backdoor neurons and preserves clean neurons by amplifying and reducing the magnitude of neuron weights, then prunes backdoor filters and high-BLC hybrid filters to balance the backdoor mitigation performance and the clean accuracy.}
\label{fra}
\end{figure}

\subsection{Proposed Method}
\paragraph{Backdoor Mitigation with MNP.}
MNP aims to prune filters that deviate from the magnitude-saliency correlation, as these filters may contain potential backdoor neurons. The weight penalty objective is designed to achieve this purpose. 
Conducting weight penalty process directly on backdoor models is enough to defend against classical attacks including BadNets \cite{guBadnetsIdentifyingVulnerabilities2017} and WaNet \cite{nguyenWaNetImperceptibleWarpingbased2021a}. However, we have empirically found that weight penalty alone is unable to thoroughly remove the backdoors injected by some advanced attacks like DFST \cite{cheng2021deep} or LIRA \cite{doan2021learnable}, as backdoor filters produced by these attacks may have smaller magnitudes, making it difficult to distinguish them from redundant filters. Therefore, we adopt the clean suppression objective to reduce the magnitude of clean neurons in the backdoor model, thereby exposing the low-magnitude backdoor filters. The combination of clean suppression and weight penalty process is enough to purify most backdoor models. To further reduce the compromise in model performance, we conduct the clean preserving process on the original backdoor model to preserve critical clean neurons, which may be mistakenly pruned and cause degradation of clean accuracy in other backdoor mitigation methods. During the process, the masks of the backdoor and redundant filters keeps almost unchanged, while the mask of a subset of clean and hybrid filters that are critical for the clean accuracy significantly increase. MNP adds the masks obtained from the clean preserving process to the masks obtained from the weight penalty process, and prunes filters with lower mask values, thus better balancing clean accuracy and backdoor mitigation performance. More discussion about the mechanism of MNP can be found in Appendix \ref{mec}.

\paragraph{Backdoor Detection with MNP.}
MNP detects backdoor models by measuring the magnitude-saliency correlation of neurons across the model. For each filter, the $l_2$-norm of its weights $\Vert\theta^{(i,j)}\Vert_2^2$ and $\textrm{CLC}(\theta, i, j)$ are chosen as the magnitude and the saliency metrics, respectively. The Spearsman's rank correlation coefficient $\rho(F,\theta)$ between $\Vert\theta^{(i,j)}\Vert_2^2$ and $\textrm{CLC}(\theta, i, j)$ can be used to measure the strength of the magnitude-saliency correlation of model $F$. 
We compare the correlations of the original model $F(\cdot;\theta)$ and the suppressed model $F(\cdot;\theta_s)$ obtained from the clean suppression process. In clean models, the unlearning process reduces the magnitude of clean neurons at approximately equal rates, so the magnitude-saliency correlation is maintained, while the correlation of backdoored models is significantly weakened. Given threshold $\delta\in [0,1]$, the set of backdoored models can be defined as follows:
\begin{equation}
    \mathcal{B}_{\delta} = \left\{F(\cdot; \theta):\frac{\rho(F, \theta_s)}{\rho(F, \theta)}\leq \delta\right\}
\end{equation}
Note that we have omitted the approximation of CLC and the normalization of the magnitude and saliency metrics. The detailed detection method can be found in Appendix \ref{det}.

\begin{algorithm}[h]
	\renewcommand{\algorithmicrequire}{\textbf{Input:}}
	\renewcommand{\algorithmicensure}{\textbf{Output:}}
\caption{Magnitude-based Pruning for Backdoor Defense (MNP)}\label{alg:mit}
\begin{algorithmic}[1]
\REQUIRE Defense data $\mathcal{D}_{d}$, backdoored model $F$ with parameters $\theta$, learning rate $\eta_1>0,\eta_2>0$, hyperparameters $\lambda$, $\mu$, max iteration number $T$, detection threshold $\delta$, pruning threshold $\epsilon \in [1,3]$
\STATE Initialize $\mathcal{M}_1 = \{[1]^{c_i}\}^L_{i=1}$, $\mathcal{M}_2 = \{[1]^{c_i}\}^L_{i=1}$, $\theta_s = \theta$
\FOR {$t=0,...,T -1$}
\STATE sample a mini-batch $\mathcal{B}=\{(\boldsymbol{x_i},y_i)\}_{i=1}^{b} \subset \mathcal{D}_{d}$
\STATE $\theta_s\leftarrow \theta_s + \eta_1\nabla_{\theta_s}\big[\mathcal{L}(F(\boldsymbol{x_i}; \theta_s), y_i) - \mu \Vert\theta_s\Vert_2^2\big]$
\ENDFOR
\STATE Compute the Spearsman's rank correlation coefficient $\rho(F, \theta), \rho(F, \theta_s)$
\REPEAT
\STATE Sample a mini-batch $\{(\boldsymbol{x_i},y_i)\}_{i=1}^{b} \subset \mathcal{D}_{d}$
\STATE $\mathcal{M}_1 \leftarrow \mathcal{M}_1 - \eta_2\nabla_{\mathcal{M}_1} \big[\mathcal{L}(F_{\mathcal{M}}(\boldsymbol{x_i}; \theta_s), y_i)
    + \lambda \sum_{i=1}^L\Vert\boldsymbol{w}_i\odot \boldsymbol{m}_i\Vert_1\big]$
\STATE $\mathcal{M}_2 \leftarrow \mathcal{M}_2 - \eta_2\nabla_{\mathcal{M}_2} \big[\mathcal{L}(F_{\mathcal{M}}(\boldsymbol{x_i}; \theta), y_i)
  - \lambda \sum_{i=1}^L\Vert\boldsymbol{w}_i\odot \boldsymbol{m}_i\Vert_1\big]$
\STATE $\mathcal{M}_1 \leftarrow \max(0, \min(\mathcal{M}_1, 1))$
\STATE $\mathcal{M}_2 \leftarrow \max(1, \min(\mathcal{M}_1, 2))$
\UNTIL {training converged}
\STATE $\hat\theta = \theta\odot \mathbb{I}\left((\mathcal{M}_1+ \mathcal{M}_2) > \epsilon \right)$ 
\ENSURE Purified model $F$ with parameters $\hat\theta$, detection result $\mathbb{I}\left(\frac{\rho(F, \theta_s)}{\rho(F, \theta)}> \delta\right)$
\end{algorithmic}
\end{algorithm}


\section{Experiments} \label{sec5}
\subsection{Experimental Setup}\label{set}
\paragraph{Attack Setup.}
We evaluate MNP against 10 challenging attacks. These include 3 static attacks: BadNets \cite{guBadnetsIdentifyingVulnerabilities2017}, Trojan \cite{liuTrojaningAttackNeural2018}, Blend \cite{chenTargetedBackdoorAttacks2017}, 2 clean label attacks: CL \cite{turnerCleanlabelBackdoorAttacks2019} and SIG \cite{barni2019new}, 2 dynamic attacks: IAB \cite{nguyenInputAwareDynamicBackdoor2020} and WaNet \cite{nguyenWaNetImperceptibleWarpingbased2021a}, 2 feature space attacks: FC \cite{shafahiPoisonFrogsTargeted2018} and DSFT \cite{cheng2021deep}, and 1 adaptive attack LIRA \cite{doan2021learnable}. 
Default settings from original papers and open-source codes are applied for most attacks, including backdoor trigger pattern and size. The backdoor label of all attacks is set to class 0, with a default poisoning rate of 10\%. Attacks are performed on CIFAR-10 with ResNet18 and a 12-class subset of ImageNet with ResNet-34. For training setups, Stochastic Gradient Descent (SGD) is used with an initial learning rate 0.1, weight decay 5e-4, momentum 0.9, batch size 128 for 200 epochs on CIFAR-10, and batch size 64 for 300 epochs on ImageNet subset. A cosine scheduler is employed to adjust the learning rate.
\paragraph{Defense Setup.}
We compare MNP with a total of 10 backdoor defense methods. These include 4 pruning-based backdoor mitigation methods: FP \cite{liuFinepruningDefendingBackdooring2018}, ANP \cite{wuAdversarialNeuronPruning2021}, CLP \cite{zhengDatafreeBackdoorRemoval2022} and RNP \cite{liReconstructiveNeuronPruning2023}, 3 non-pruning mitigation methods: NC \cite{wangNeuralCleanseIdentifying2019}, NAD \cite{liNeuralAttentionDistillation2020} and I-BAU \cite{zengAdversarialUnlearningBackdoors2022}, as well as 4 detection methods: NC, AC \cite{chenDetectingBackdoorAttacks2018}, SC \cite{DBLP:conf/nips/Tran0M18} and STRIP \cite{gaoSTRIPDefenceTrojan2020}. All the defenses share limited access to 1\% benign training data except for CLP. Hyperparameters for these defenses are adjusted based on open-source codes to obtain best performance against different attacks. For MNP, we set hyperparameter $\mu$ as 0.001, $\lambda$ as 0.0005, suppression epochs $T$ as 20, learning rates $\eta_1$ as 0.01 and $\eta_2$ as 0.1. The detection threshold $\delta$ is set as $0.2$, and the pruning threshold $\epsilon$ is dynamically adjusted to prune 30 neurons for each backdoor model.
\paragraph{Evaluation Metric.}
We adopt two metrics for evaluating backdoor mitigation performance: 1) Clean Accuracy (CA), which is the model's accuracy on clean test data; 2) Attack Success Rate (ASR), which is the model's accuracy on backdoored test data. We adopt 3) Detection Rate (DR) to evaluate detection performance, which is the accuracy of the defense in identifying backdoor models.

\begin{table*}[!t]
\caption{Comparison with the state-of-the-art defenses on \textbf{CIFAR-10} dataset with 1\% benign data on ResNet18 (\%).}
\centering
\renewcommand\arraystretch{1.5} 
\resizebox{\textwidth}{!}{%
\begin{tabular}{@{}c|c|c|c|c|c|c|c|c|cc@{}}
\toprule
\multirow{2}{*}{Attack} &
  \textbf{Backdoored} &
  \textbf{FP} &
  \textbf{ANP} &
  \textbf{CLP} &
  \textbf{RNP} &
  \textbf{NC} &
  \textbf{NAD} &
  \textbf{I-BAU} &
  \textbf{MNP(Ours)} \\
 &
  CA/ ASR &
  CA/ ASR &
  CA/ ASR &
  CA/ ASR &
  CA/ ASR &
  CA/ ASR &
  CA/ ASR &
  CA/ ASR &
  CA/ ASR \\ \hline 
  BadNets &
  92.43  /  100.00 &
  81.45 / 25.31 &
  90.27 / 1.12 &
  91.54 / 1.34 &
  91.09 / 0.54 &
  89.32 / 5.54 &
  89.57 / 1.10&
  90.19 / 12.73&
  \textbf{92.11} / \textbf{0.47} \\

Trojan &
  92.68 / 100.00 &
  82.63 / 62.57 &
  90.78 / 1.31 &
  91.16 / 2.87 &
  91.95 / 2.03 &
  90.91 / 52.72 &
  86.73 / 5.74 &
  90.35 / 10.55 &
  \textbf{92.24} / \textbf{0.92} \\
Blend &
  92.15 / 99.99 &
  83.26 / 76.44 &
  90.82 / 0.90 &
  90.61 / 1.67 &
  91.53 / 1.33 &
  \textbf{91.84} / 84.31 &
  89.68 / 13.24 &
  89.94 / 2.24 &
  91.79 / \textbf{0.87} \\
CL &
  91.55 / 98.93 &
  81.38 / 36.42 &
  89.96 / 5.47 &
  89.51 / 1.54 &
  90.05 / 0.75 &
  90.13 / 5.66 &
  86.74 / 15.18 &
  87.75 / 20.12 &
  \textbf{91.18} / \textbf{0.62}
  \\
SIG &
  93.52 / 99.14 &
  88.17 / 23.56 &
  91.57 / 5.39 &
  90.14 / 10.35 &
  90.63 / \textbf{0.89} &
  90.50 / 90.15 &
  91.37 / 3.46 &
  86.71 / 25.62 &
  \textbf{93.50} / 1.32
  \\
IAB &
  94.60 / 99.66 &
  85.42 / 38.95 &
  93.67 / 1.52 &
  93.20 / 4.71 &
  93.15 / 2.34 &
  94.19 / 97.98 &
  89.94 / 12.16&
  87.68 / 18.34 &
  \textbf{94.45} / \textbf{0.48} &
  \\
WaNet &
  92.12 / 98.81 &
  80.53 / 69.74 &
  91.06 / 8.89 &
  90.58 / 6.43 &
  91.60 / 4.02 &
  91.03 / 96.50 &
  83.32 / 13.18 &
  88.91 / 25.48 &
  \textbf{91.75} / \textbf{3.64} &
  \\
FC &
  93.61 / 100.00 &
  88.92 / 98.04  &
  85.95 / 77.42 &
  80.21 / 65.87 &
  89.39 / \textbf{1.55} &
  \textbf{92.41} / 99.78 &
  90.14 / 30.37 &
  85.79 / 18.22 &
  91.23 / 1.87 &
  \\
DFST &
  95.50 / 100.00 &
  85.53 / 80.76 &
  91.24 / 19.80 &
  87.45 / 58.82 &
  92.15 / 25.67 &
  93.51 / 99.02 &
  87.43 / \textbf{15.70} &
  85.22 / 26.84 &
  \textbf{94.79} / 20.40 &
  \\
LIRA &
  91.20 / 97.78 &
  86.64 / 90.52 &
  84.17 / 21.25 &
  80.38 / 60.56 &
  89.76 / 18.62 &
  \textbf{90.08} / 97.30 &
  86.52 / 30.15 &
  84.33 / 57.09 &
  89.29 / \textbf{9.72} &
  \\ \midrule
Average &
  94.42 / 98.83 &
  84.39 / 60.23 &
  89.95 / 14.31 &
  88.68 / 21.41 &
  91.39 / 7.60 &
  91.46 / 72.90 &
  88.14 / 14.03 &
  88.69 / 21.72 &
  \textbf{92.23} / \textbf{4.03}\\
\bottomrule
\end{tabular}%
\label{table1}
}
\end{table*}

\begin{table*}[!t]
\caption{Comparison with the state-of-the-art defenses on \textbf{ImageNet subset} dataset with 1\% benign data on ResNet34 (\%).}
\centering
\renewcommand\arraystretch{1.5} 
\resizebox{\textwidth}{!}{%
\begin{tabular}{@{}c|c|c|c|c|c|c|c|c|cc@{}}
\toprule
\multirow{2}{*}{Attack} &
  \textbf{Backdoored} &
  \textbf{FP} &
  \textbf{ANP} &
  \textbf{CLP} &
  \textbf{RNP} &
  \textbf{NC} &
  \textbf{NAD} &
  \textbf{I-BAU} &
  \textbf{MNP(Ours)} \\
 &
  CA/ ASR &
  CA/ ASR &
  CA/ ASR &
  CA/ ASR &
  CA/ ASR &
  CA/ ASR &
  CA/ ASR &
  CA/ ASR &
  CA/ ASR \\ \hline 
BadNets &
  91.89  /  100.00 &
  81.43 / 93.98 &
  89.18 / 6.34 &
  85.52 / 9.81 &
  90.16 / 1.54 &
  83.80 / 10.27 &
  81.45 / 9.83&
  80.72 / 20.95&
  \textbf{92.21} / \textbf{0.87} \\

Trojan &
  91.50 / 99.87 &
  81.17 / 90.93 &
  \textbf{90.31} / 1.78 &
  85.98 / 7.51 &
  89.06 / 1.51 &
  85.53 / 80.98 &
  83.05 / 5.74 &
  84.36 / 10.55 &
  90.03 / \textbf{1.08} \\
Blend &
  89.24 / 100.00 &
  79.05 / 84.41 &
  82.18 / 9.20 &
  82.69 / 3.31 &
  87.20 / 4.37 &
  83.86 / 97.93 &
  79.03 / 12.71 &
  80.21 / 19.35 &
  \textbf{89.04} / \textbf{2.56} \\
CL &
  88.91 / 90.32 &
  81.44 / 84.73 &
  80.75 / 12.81 &
  84.30 / 8.07 &
  87.13 / 5.05 &
  80.66 / 30.21 &
  83.17 / 24.76 &
  84.06 / 11.57 &
  \textbf{86.92} / \textbf{3.41}
  \\
WaNet &
  88.69 / 95.48 &
  82.17 / 90.32 &
  81.62 / 14.73 &
  79.14 / 21.56 &
  84.91 / 13.68 &
  83.25 / 94.56 &
  81.40 / 33.15 &
  80.82 / 44.68 &
  \textbf{86.85} / \textbf{8.42} &
  \\
FC &
  87.58 / 94.36 &
  79.45 / 87.42  &
  73.81 / 53.69 &
  75.33 / 77.92 &
  82.49 / \textbf{10.32} &
  86.60 / 92.68 &
  81.65 / 56.40 &
  79.36 / 38.35 &
  \textbf{87.04} / 10.63 &
  \\ \midrule
Average &
  89.64 / 96.67 &
  80.79 / 88.63 &
  82.98 / 16.43 &
  82.16 / 21.36 &
  86.83 / 6.08 &
  83.95 / 67.77 &
  81.63 / 23.77 &
  87.53 / 24.24 &
  \textbf{88.68} / \textbf{4.10}\\
\bottomrule
\end{tabular}%
\label{table2}
}
\end{table*}

\subsection{Main Defense Results}
\paragraph{Results on CIFAR-10.} Table \ref{table1} presents the defense performance of 8 backdoor mitigation methods against 10 backdoor attacks on CIFAR-10. MNP outperforms other defense methods by cutting the average ASR down to 4.03\% with a slight drop on CA (2.19\% on average). In comparison, the state-of-the-art methods ANP, RNP, NAD and I-BAU reduce the average ASR to 14.31\%, 7.60\%, 14.03\%, and 21.72\%, respectively. An important observation is that each defense method has its limits. For instance, NAD achieves the lowest ASR (15.70\%) on DFST but is outperformed by MNP on most other attacks. MNP shows weakness against SIG and FC, yet it has the best overall performance and highest clean accuracy in most settings. Nearly all defenses struggle against DFST and LIRA, suggesting the need for more sophisticated mechanisms to counter advanced attacks.

\paragraph{Results on ImageNet Subset.}
Table \ref{table2} presents the defense performance of 8 backdoor mitigation methods against 6 backdoor attacks on CIFAR-10. Note that some attacks are not conducted for ImageNet subset because of difficulties in reproduction on high-resolution dataset. Pruning larger models trained on high-resolution dataset poses a greater challenge due to the difficulty in locating backdoor neurons. However, MNP can effectively defend against existing attacks and still demonstrates an advantage in maintaining clean accuracy. For example, MNP can reduce the ASR of BadNet to 0.87\%, even slightly improving the clean accuracy. MNP also generally outperforms most state-of-the-art defenses, except for a slightly lower clean accuracy (90.03\%/90.31\%) on Trojan compared to ANP and a slightly weaker ASR (10.63\%/10.32\%) on FC compared to RNP. Nevertheless, it is evident that MNP can better balance clean accuracy and ASR overall, as it cuts the average ASR down to 4.10\% with only <1\% decline in CA.
\paragraph{Detection Performance.}
The clean suppression process of MNP can be use to detect backdoor models by the strength of magnitude-saliency correlation, and the experimental results can be found in \ref{tab:det} in Appendix \ref{add}. MNP is able to detect backdoor models generated by six attacks and achieve a 98.00\% DR, outperforming NC (51.33\%), AC (89.75\%), SC (83.08\%), and STRIP (80.67\%).
\subsection{Ablation Studies}

\begin{figure}
    \begin{minipage}[b]{0.46\textwidth}
        \includegraphics[width=\textwidth]{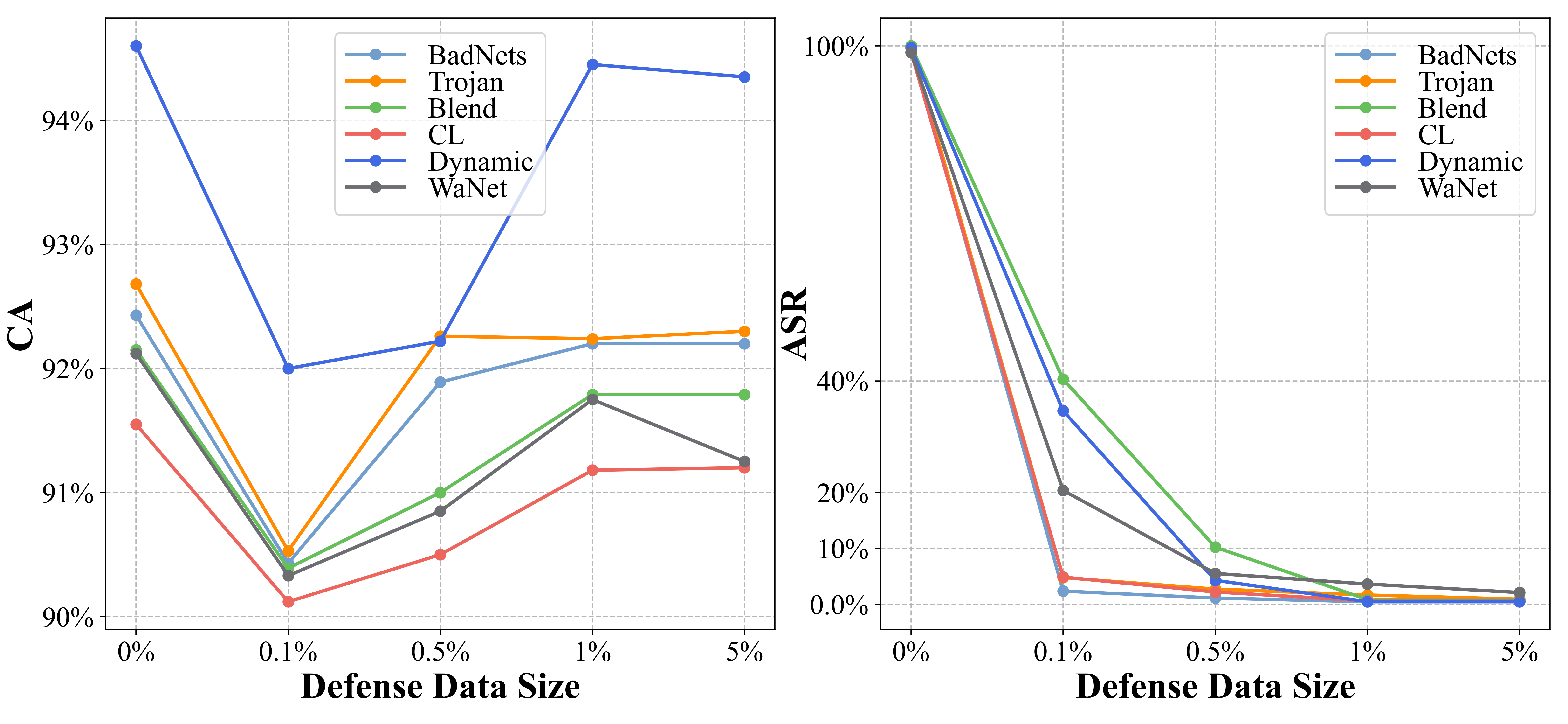}
            \vspace{-0.5em}
        \caption{Defense performance of MNP with different defense data size against BadNets}\label{hyp}
    \end{minipage}
    \begin{minipage}[b]{0.46\textwidth}
        \includegraphics[width=\textwidth]{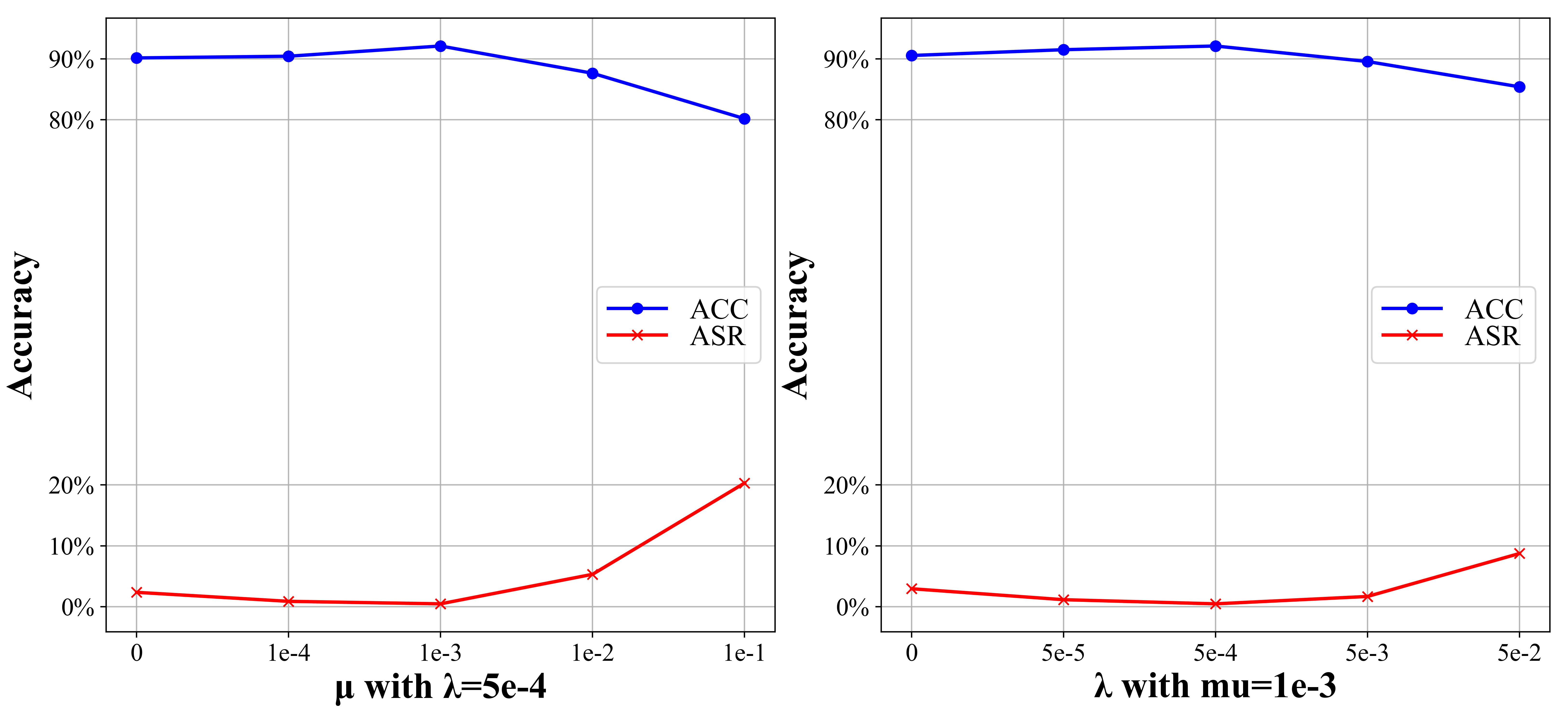}
            \vspace{-0.5em}
        \caption{Defense performance of MNP with different hyperparameter settings}\label{datasize}
    \end{minipage}
\end{figure}

\paragraph{Impact of the Defense Data Size.}
In this part, we evaluate the impact of defense data size on the performance of MNP with a backdoored ResNet18 attacked by BadNets. We use 0.1\%(50), 0.5\% (250), 1\% (500) and 5\% (2500) images from the CIFAR-10 training set for defense, respectively. Results in Fig \ref{datasize} show that as defense data size increases, MNP demonstrates better backdoor mitigation performance. In some cases, MNP only needs 0.1\% of clean samples to reduce the ASR of several attacks to < 5\%, highlighting its potential to identify backdoor neurons in few-shot settings. The impact of defense data size is more pronounced on CA. As the number of clean samples grows, the distribution of the defense set gets closer to the original dataset, allowing MNP to more accurately preserve clean neurons. Overall, MNP can effectively eliminate backdoor neurons under few-shot settings, and 1\% of clean samples is sufficient for MNP to limit the degradation of CA to < 3\%.

\begin{table}[!t]
\caption{Defense performance of MNP against BadNets attack on CIFAR-10 with different poisoning rates and A2O/A2A settings}
\centering
\renewcommand\arraystretch{1.1}
\setlength{\tabcolsep}{4pt}
\scalebox{0.674}{%
\begin{tabular}{cc|cc|cc|cc|cc|cc}
\toprule
\multicolumn{2}{c|}{Poisoning Ratio $\rightarrow$} &
\multicolumn{2}{c|}{0.05\%(250)} &
\multicolumn{2}{c|}{1\%(500)} &
\multicolumn{2}{c|}{5\%(2500)} &
\multicolumn{2}{c|}{10\%(5000)} &
\multicolumn{2}{c}{20\%(10000)} \\
\multicolumn{2}{c|}{ATTACK $\downarrow$} &
No Defense &
MNP &
No Defense &
MNP &
No Defense &
MNP &
No Defense &
MNP &
No Defense &
MNP \\ \midrule
\multirow{2}{*}{BadNets-A2O} & CA & 92.35 & 85.76 & 92.47 & 88.29 & 92.60 & 92.15 & 92.43 & 92.11 & 89.97 & 88.31 \\
& ASR & 99.05 & 10.40  & 98.93 & 3.26  &100.00 & 0.39  & 100.00 & 0.47  & 100.00 & 1.28  \\ \midrule
\multirow{2}{*}{BadNets-A2A}     & CA & 93.05 & 92.02 & 92.97 & 92.16 & 92.44 & 91.89 & 92.65 & 92.07 & 86.31 & 84.12 \\
& ASR & 63.51 & 44.38  & 79.52 & 8.75  &90.45 & 0.78  & 91.38 & 0.97  & 94.56 & 0.44\\ \bottomrule
\end{tabular}%
\label{table5}
}
\vspace{-1em}
\end{table}

\paragraph{Performance against Different Poisoning Rate and All-to-all attack.}
We conducted experiments on MNP with different poisoning rates on the CIFAR-10. We have also tested the effectiveness of MNP against the all-to-all attack, where the target label of the backdoored sample is set to one plus the original label ($S(y) = y+1$). Experimental results are shown in Table 5. For attacks with poisoning rate $\geq$ 5\%, MNP can effectively reduce the ASR to approximately 1\% without causing significant CA decrease. For attacks with poisoning rates $\leq$ 1\%, MNP can still maintain a certain level of defense performance. Additionally, MNP can effectively defend against all-to-all attacks with poisoning rates $\geq$ 5\%, reducing the ASR to 1\% with < 2\% CA degradation.

\paragraph{Impact of hyperparameters.}
The most critical hyperparameters in MNP are $\lambda$ and $\mu$, as higher values of $\lambda$ or $\mu$ cause the optimization process to be dominated by the magnitude of neurons rather than the classification loss of the model. We kept other hyperparameters the same as in \ref{set} and adjusted $\mu$ to values of $0$, $1e^{-4}$, $1e^{-3}$, $1e^{-2}$, and $1e^{-1}$, and lambda to values of $0$, $5e^{-5}$, $5e^{-4}$, $5e^{-3}$, and $5e^{-2}$. The results in Fig \ref{hyp} indicate that MNP achieves the best defense performance when $\mu$ and $\lambda$ are set to proper, smaller values. A larger $\mu$ simultaneously reduces the magnitude of both backdoor and clean neurons and make the clean suppression process ineffective. Similarly, a larger $\lambda$ makes MNP ignore the classification loss, which prevents it from effectively preserving clean neurons. Overall, setting $\lambda$ and $\mu$ to smaller values helps maintain stable performance, while appropriate larger $\lambda$ and $\mu$ helps achieve the best performance.

\section{Conclusion}
This paper proposes Magnitude-based Neuron Pruning (MNP), a novel method to detect and mitigate backdoor attacks. The core idea of MNP is to manipulate the magnitude of backdoor and clean neurons through three process named clean suppression, weight penalty and clean preserving. Clean suppression reduces the magnitude of clean neurons to expose and identify backdoor behavior. Weight penalty eliminates neurons with large magnitude but less contribution to the clean task, i.e., potential backdoor neurons. Clean preserving aims to increase the magnitude of critical clean neurons to avoid them from being pruned. The empirical success of MNP across a range of backdoor attack scenarios highlights the potential of neuron magnitude for backdoor defense.

\clearpage
{
    \small
    \bibliography{MNP}
    \bibliographystyle{plainnat}
}
\newpage
\appendix
\section{Theoretical Analysis and Mechanism of MNP} \label{apdix_B}
\subsection{Min-Max Optimization for Backdoor Defense}\label{minmax}
A number of recent works \cite{wuAdversarialNeuronPruning2021,liReconstructiveNeuronPruning2023,chaiOneshotNeuralBackdoor2022,zhuEnhancingFineTuningBased2023} on backdoor mitigation have employed a min-max optimization process to expose backdoor neurons. This paradigm can be reinterpreted from the perspective of neuron magnitude.

ANP \cite{wuAdversarialNeuronPruning2021}, one of the state-of-the-art backdoor mitigation methods, adversarially perturbs the magnitude of each filter to maximize the clean loss of the model. 
The adversarial perturbation can possibly improve the magnitude of the backdoor and hybrid filters, thus amplifying the backdoor activation and trigger the backdoor behavior. 
Then, ANP recovers the model performance by optimizing masks for each filter, during which process the masks of filters contributing most to the backdoor behavior are reduced to zero, thus can be identified and pruned.

Another state-of-the-art method RNP \cite{liReconstructiveNeuronPruning2023} unlearns the model to maximize the clean loss. The difference is that the unlearning process is conducted at the neuron level.
During unlearning, the magnitude of clean neurons is reduced, and the magnitude of backdoor neurons may also increase. 
Then, RNP apply a similar filter-level recovering process to recover the clean accuracy and expose the backdoor filters. 
RNP is empirically more effective than ANP
, as the unlearning process possibly reduce the magnitude of most clean neurons
, causing the magnitude of all backdoor neurons relatively increases. Since the clean neurons are neutralized, some hybrid filters may also transform into backdoor filters. 
In this way, the majority of the backdoor-associated filters are exposed. 
In contrast, the adversarial perturbation in ANP only amplifies the magnitude of a subset of backdoor filters, some low-contribution backdoor filters and hybrid filters are generally ignored.

\subsection{More Understanding of MNP}\label{mec}
\paragraph{Why clean suppression is needed before weight penalty.}
We assume that backdoor or hybrid filters contain backdoor neurons and have larger $l_p$-norms that deviate from the magnitude-saliency correlation. 
We find this assumption holds strongly for highly regularized models, and directly conducting the weight penalty process for these models is enough to erase the injected backdoor.
However, for less regularized models or models not fully converged, the magnitude-saliency correlation is not as obvious as the highly regularized models.
In this case, the weight penalty process is not enough to thoroughly remove the inject backdoor. 
By applying the clean suppression process before weight penalty, we can reduce the magnitude of most clean neurons and expose more backdoor-associated filters, 
leading to better backdoor mitigation performance.

\paragraph{Why clean suppression not adversarial perturbation.}
As is discussed in Section \ref{minmax}, the adversarial perturbation, whether conducted at filter level or neuron level, 
can only amplifying the magnitude a subset of high-contribution backdoor filters, since the classification loss can be effectively increased by perturbing a small number of backdoor neurons. 
In contrast, the clean suppression process reduces the magnitude of most clean neurons, thus the magnitude of all backdoor neurons relatively increases and more backdoor-associated filters can be exposed.

\paragraph{How can clean preserving help in reducing compromise of clean accuracy.}
Backdoor neurons exist across different layers of the infected model. Some hybrid filters (for example, filters in the first convolution layer) may contain clean neurons critical for the model performance, 
which can be possibly neutralized by the clean suppression process. 
Although the weight penalty process can expose most backdoor and hybrid filters, it can not completely recover the clean accuracy. 
In other word, some hybrid filters that are critical for the clean task may be pruned. 
To reduce the compromise of clean accuracy, we apply the clean preserving process to preserve critical clean neurons. 
The clean preserving process is performed at the filter level. 
During the process, the masks of the backdoor and redundant filters keeps almost unchanged, 
while the mask of a subset of clean and hybrid filters that are critical for the clean accuracy significantly increase. 
We add the masks obtained from the clean preserving process to the masks obtained from the weight penalty process, and prune filters with lower mask values, 
thus better balancing clean accuracy and backdoor mitigation performance. 
The pruning strategy of MNP is different from most of the previous methods \cite{wuAdversarialNeuronPruning2021,liReconstructiveNeuronPruning2023}, which prune filters directly by their contribution to the backdoor task. For example, if a hybrid filter contributes significantly to both the clean task and the backdoor task, it may be directly pruned in these methods. In contrast, in MNP, as the clean preserving process increase the mask of that filter, the priority of pruning it is lower than pruning filters that reduce the backdoor contribution but do not affect the clean accuracy, i.e., purely backdoor neurons.

\subsection{Detect Backdoor Models with MNP} \label{det}
Based on the assumptions in Section \ref{ass}, the magnitude of a neuron is positively correlated with its contribution to the prediction results. In benign models, the magnitude-CLC correlation is approximately equal to the magnitude-contribution correlation. In backdoor models, the magnitude-CLC correlation should be weaker than the actual magnitude-contribution correlation, for that CLC only partially reflects the contribution of backdoor and hybrid neurons, as mentioned in Section \ref{obs}.
Intuitively, we can detect backdoor models by measuring the strength of the correlation between neuron magnitude and saliency. However, directly computing CLC is computationally expensive, for it requires pruning each filter $\theta^{(i,j)}$ and measuring the change in loss. The clean loss $\mathcal{L}_{cl}$ can be denoted as follows:
\begin{equation}
    \mathcal{L}_{cl}(\theta)=\mathbb{E}_{(\boldsymbol{x}, y) \in \mathcal{D}_t}\mathcal{L}(F(\boldsymbol{x}; \theta), y)
\end{equation}
We can approximate CLC in the vicinity of $\theta$ by the first-order Taylor expansion of the clean loss:
\begin{equation}\label{tay}
    \mathcal{L}_{cl}(\theta) = \mathcal{L}_{cl}(\theta|{\theta^{(i,j)}=0}) + \frac{\delta\mathcal{L}_{cl}}{\delta \theta^{(i,j)}}\theta^{(i,j)}+O(\Vert\theta^{(i,j)}\Vert_p^2)
\end{equation}
We can neglect the first-order remainder to avoid computational difficulties, since the widely-used ReLU activation function encourages a smaller second-order term. By substituting Eq. \ref{tay} into Eq. \ref{clc} and ignoring the remainder, we have:
\begin{equation}\label{sal}
\widetilde{\textrm{CLC}}(\theta,i,j) = 
\mathcal{L}_{cl}(\theta) - \frac{\delta\mathcal{L}_{cl}}{\delta \theta^{(i,j)}}\theta^{(i,j)} - \mathcal{L}_{cl}(\theta) = - \frac{\delta\mathcal{L}_{cl}}{\delta \theta^{(i,j)}}\theta^{(i,j)},
\end{equation}
which is computationally friendly, as the gradient $\frac{\delta\mathcal{L}_{cl}}{\delta \theta^{(i,j)}}$ can be easily computed through backpropagation. Another problem is that the scale of neuron magnitude and saliency varies across different layers of DNNs. Thus we apply a simple layer-wise $l_2$-normalization to conduct rescaling across layers. The magnitude and saliency metrics for each filter are defined as follows:
\begin{align}
m_{ij} &= \frac{\Vert\theta^{(i,j)}\Vert_p}{\sqrt{\sum_j(\Vert\theta^{(i,j)}\Vert_p)^2}}\\
s_{ij} &= \frac{|\widetilde{\textrm{CLC}}(\theta,i,j)|}{\sqrt{\sum_j(\widetilde{\textrm{CLC}}(\theta,i,j)^2}}
\end{align}
Note that it's necessary to use the absolute value of CLC as the saliency metric, since CLC can not reflect the uncertainty of model prediction brought by pruning a single filter, as discussed in model compression studies \cite{DBLP:conf/cvpr/MolchanovMTFK19,DBLP:conf/iclr/MolchanovTKAK17}. For each model $F$, we compute the Spearman's rank correlation coefficient $\rho_F$ between the magnitude metric $m_{ij}$ and the saliency metric $s_{ij}$. It is clear that under the same training settings, the coefficient for backdoor models is significantly lower than clean metrics. 

However, the defender lacks information about the training process and can not compare the untrustworthy model with the corresponding benign model. We instead consider tune the model with the clean suppression objectives to perturb the magnitude of clean neurons. 
In clean models, the unlearning process reduces the magnitude of clean neurons at approximately equal rates, so the magnitude-saliency correlation is maintained, while the correlation of backdoored models is significantly weakened.
The idea is that the unlearning process has different effects on backdoor and clean neurons, making the hybrid filters and backdoor filters further deviate from the magnitude-saliency correlation, which is also implied in \cite{liReconstructiveNeuronPruning2023}.
\subsection{Limitations of MNP} \label{limitations}
Despite the promising results of our Magnitude-based Neuron Pruning (MNP) method, there are several limitations that warrant attention:
\begin{enumerate}
    \item \textbf{Clean data acquisition.} Although MNP requires only a modest amount of clean data to achieve effective backdoor defense, the availability of clean data remains a significant limitation. Without access to a reliable source of clean data, the application of our method may be restricted. Future work should focus on enhancing MNP to operate with even less clean data or develop techniques that can identify and utilize clean data within a poisoned dataset.
    \item \textbf{Defense against low poisoning rate attacks.} MNP shows robust performance against a variety of backdoor attacks but faces challenges when the poisoning rate is exceedingly low (e.g., $\leq 1\%$). In such cases, backdoor neurons are more adept at blending with clean neurons, making them harder to detect and prune accurately. Addressing this limitation will involve refining our detection algorithms to be more sensitive to subtle anomalies in neuron behavior.
    \item \textbf{Theoretical guarantees against advanced attacks.} Our study is grounded in observations and regularization assumptions, which means that we cannot theoretically guarantee that MNP can defend against all sophisticated attacks. The evolving nature of adversarial strategies necessitates continuous research to strengthen the theoretical underpinnings of our method and ensure it remains effective against future threats.
\end{enumerate}


\section{Additional Experimental Results}\label{add}
\begin{table*}[!tbp]
\small
\centering
\caption{Comparison of DR with the 4 different detection methods against 6 existing attacks (\%)}
\begin{adjustbox}{width=0.8\linewidth}
    \begin{tabular}{c|cccccc|c}
    \toprule
    Defense & BadNets & Trojan & Blend & CL & IAB & WaNet &\ Avg \\ \hline
    NC & 100.00 & 100.00 & 20.50 & 52.50 & 9.50 & 25.50 &  51.33 \\
    AC & 100.00 & 100.00 & 91.50 & 84.50 & 75.50 & 87.00 &  89.75 \\
    SC  & 100.00 & 100.00 & 95.50 & 55.00 & 57.50 & 90.50 &  83.08 \\ 
    STRIP  & 100.00 & 100.00 & 83.50 & 50.00 & 50.50 & 100.00 & 80.67 \\ 
    MNP(Ours)  & 100.00 & 100.00 & 100.00 & 95.50 & 92.50 & 100.00 & 98.00\\ 
    \bottomrule
    \end{tabular} 
\end{adjustbox}
\vskip -0.1in
 \label{tab:det}
\end{table*}

\subsection{Run-time Analysis} 
We record the running time of MNP on a RTX 3090Ti GPU with 500 CIFAR-10 samples and a ResNet18 model attacked by BadNets. It costs MNP 1 minutes and 27 seconds to reduce the ASR to 0.47, closer to FP (30 seconds), ANP (45 seconds) and RNP (1 minutes and 11 seconds). The computing cost is acceptable compared to retraining the model from scratch (more than 1 hour).

\end{document}